\documentclass[10pt,twocolumn,letterpaper]{article}

\usepackage{cvpr}              
\usepackage{graphicx} 
\usepackage{booktabs} 
\usepackage{amsmath}  

\usepackage{marvosym}
\usepackage{pifont}
\newcommand{\cmark}{\ding{51}}
\newcommand{\xmark}{\ding{55}}

\usepackage{xcolor}   
\usepackage{colortbl} 
\usepackage{appendix} %
\usepackage{amsthm}
\usepackage{algorithm}
\usepackage{float}
\usepackage{caption}
\usepackage{algorithmic}  
\theoremstyle{plain} 

\usepackage{bm}
\usepackage{multirow}
\usepackage{amsmath}
\usepackage{mathtools}

\usepackage{xr} 
\newcommand\nnfootnote[1]{%
  \renewcommand\thefootnote{}\footnote{#1}%
  \addtocounter{footnote}{-1}%
}

\definecolor{cvprblue}{rgb}{0.21,0.49,0.74}
\usepackage[pagebackref,breaklinks,colorlinks,allcolors=cvprblue]{hyperref}

\def\paperID{****} 

\def\confName{CVPR}
\def\confYear{2026}

\title{Efficient Long-Context Modeling in Diffusion Language Models \\
via Block Approximate Sparse Attention}

\author{
Wenhu Zhang\textsuperscript{1} \quad
Yiming Wu\textsuperscript{2} \quad
Huanyu Wang\textsuperscript{3} \quad
Yaoyang Liu\textsuperscript{1} \quad
Huanzhang Dou\textsuperscript{3} \\
Senqiao Yang\textsuperscript{4} \quad
Sitong Wu\textsuperscript{4} \quad
Hanbin Zhao\textsuperscript{3} \quad
Jiaya Jia\textsuperscript{1\Letter} \\[0.3em]
\textsuperscript{1}The Hong Kong University of Science and Technology \quad
\textsuperscript{2}The University of Hong Kong\\
\textsuperscript{3}Zhejiang University \quad
\textsuperscript{4}The Chinese University of Hong Kong\\[0.3em]
}

\newcommand{\OUR}{Block Approximate Sparse Attention}
\newcommand{\OURshort}{BA-Att}

\begin{document}
\maketitle
\begin{abstract}
Diffusion Language Models (DLMs) enable globally coherent, bidirectional, and controllable text generation, offering advantages over traditional autoregressive LLMs, while scaling to ultra-long sequences remains costly. Many existing block-sparse attention methods select blocks by fixed sampling patterns over the high-resolution attention space, e.g. tail-regions or anti-diagonal stripes. Such prior-driven sampling can miss salient tokens and introduce instability under distribution shifts. 
In this paper, we propose the \OUR\ framework (\OURshort) with block-wise pre-downsampled operation, which identifies informative regions within a compact downsampled spaces, avoiding reliance on brittle positional priors. 
To analyze its theoretical behavior, we define an oracle post-downsample attention map and formalize the approximation error between pre- and post-downsample schemes. 
Based on this insight, we introduce a lightweight norm-sorting module and a covariance-compensated correction that approximates full covariance using diagonal QK variances, reducing large computational complexity.
Extensive experiments show that our operator achieves up to \textbf{6.95× acceleration} over FlashAttention in attention computation, and maintains \textbf{near full-attention performance} at 50\% sparsity 
across language models, multi-modal language models, and video generation models, demonstrating strong efficiency and generalization.

\end{abstract}


\nnfootnote{
\hspace{-2em}{Code will be publicly available at:
\url{https://github.com/JIA-Lab-research/Block-Approximate-Sparse-Attention}.} \\
\hspace{2em}\Letter~{Corresponding Author.}
}

\vspace{-3mm}
\section{Introduction}
\label{sec:intro}


\begin{table*}[ht]
    \small
    \setlength{\tabcolsep}{15pt}
    \centering
    \caption{Comparison of sparse attention selection strategies. 
    \textbf{Pattern-Agnostic}: no reliance on fixed structural priors (e.g., sink, vertical, diagonal). 
    \textbf{Downsampled Search}: pattern selection is performed in a downsampled token or block space. 
    \textbf{Search Space Coverage}: fraction of the full $Q \times K$ attention matrix examined during selection. 
    \textbf{Search Complexity}: asymptotic cost of the selection stage w.r.t. sequence length $L$, with block size $B$ (typically 64 or 128 tokens) and stride $s$ (typically 8 or 16). }
  \label{tab:intro_search_comp}
  \begin{tabular}{lccccc}
    \toprule
    \textbf{Method} & \textbf{Pattern-Agnostic}  & \textbf{Training-Free}  & \textbf{Search Space Coverage} & \textbf{Search Complexity} \\
    \midrule
    MInference~\citep{minference}    & \xmark & \cmark   & $< 5\%$      & $O(L B)$ \\
    FlexPrefill~\citep{flexprefill}  & \xmark & \cmark   & $< 5\%$      & $O(L B)$ \\
    XAttention~\citep{xatt}          & \xmark & \cmark   & $12.5\%$     & $O(L^2 / s)$ \\
    SeerAttention~\citep{seeratt}    & \cmark   & \xmark & $100\%$      & $O((L/B)^2)$ \\
    \rowcolor{cyan!15}
    \textbf{Ours}                    & \cmark   & \cmark  & $100\%$      & $O((L/B)^2)$ \\
    \bottomrule
  \end{tabular}
\end{table*}

Diffusion Language Models (DLMs), \emph{a.k.a.} Masked Diffusion Models, are rapidly emerging as a non-autoregressive alternative to traditional autoregressive LLMs. Compared to models such as GPT or LLaMA, DLMs offer several potential advantages in generation dynamics, contextual understanding, and controllability. Specifically, DLMs formulate text generation as a gradual denoising process that iteratively refines a noisy linguistic representation into a clean sentence. This paradigm enables globally coherent generation with improved long-range and bidirectional context modeling, positioning DLMs as a key direction for next-generation generative language modeling.

However, scaling DLMs to ultra-long contexts poses significant computational challenges. The introduction of sparse attention has been a key step toward improving efficiency, allowing models to focus computation on structurally or semantically salient regions. Two typical routines are proposed to solve this problem. The first is learning gating modules to predict importance. \eg, SeerAttention~\citep{seeratt} proposes a mechanism to learn and operate the downsampled block spaces. However, these methods require additional training, limiting real world deployment.
Second, motivated by empirical observations, a series of methods attempt to design training-free strategies, such as the A-Shape pattern from StreamingLLM~\cite{xiaoefficient}, the Vertical-Slash mechanisms from MInference~\citep{minference}, query-specific patterns in FlexPrefill~\citep{flexprefill}, and the Antidiagonal scoring strategy from XAttention~\citep{xatt}. 
However, most training-free sparse attention relying on fixed positional priors to sample from the high-resolution attention space would miss salient tokens. 
%
This raises a fundamental question: \emph{Can we design a \textbf{training-free} sparse attention method that operates effectively in the \textbf{downsampled attention space}, while preserving both efficiency and robustness without any fine-tuning ?}

In this paper, we propose \OUR (\textbf{\OURshort}), a block-wise downsampling framework. Our method introduces a pre-downsampling stage that evaluates the information quality of each block directly in the downsampled attention space, thereby avoiding reliance on heuristic positional priors.
To ground the method in theory, we analyze the ideal upper bound and formalize the approximation error between the pre-downsample and post-downsample schemes, providing a principled characterization of the efficiency–fidelity trade-off inherent in block-sparse attention. This formulation further allows us to define an oracle attention map induced by the post-downsampling computation. From the high-level comparisons in Table~\ref{tab:intro_search_comp}, BA-Att achieves full coverage in a downsampled space without any post-training, while others either depend on fixed structural assumptions or suffer from limited search coverage.

Building upon the theoretical analysis, we show that the approximation error grows with the intra-block norm range of QK pairs; in other words, large norm disparities within a block lead to dispersed attention patterns and degrade performance. To address this issue, we introduce a lightweight norm-sorting module with complexity $O(L \times D)$, which reorders tokens by their norm magnitude to reduce approximation difficulty and cluster high-activation regions. Based on the resulting error formulation, we further derive a covariance-compensated correction scheme that improves accuracy without sacrificing efficiency. Specifically, we approximate the full covariance term using the diagonal combination of QK variances, reducing the overall attention complexity from $O(L \times D^2)$ to $O(L \times D)$.

We conduct extensive experiments to validate the efficiency and effectiveness. Under a \emph{\textbf{sequence length of 128K}}, our method achieves a 6.95× speedup over Flash-Attention. 
When integrated into \emph{\textbf{diverse DLMs}}, our method consistently matches the performance of full attention at 50\% sparsity.
\emph{\textbf{In video generation}}, our method surpasses SOTA video generation methods, clearly demonstrating the superior fidelity–efficiency trade-off.
The contributions are summarized as follows:



\begin{itemize}
\item \textbf{Pre-downsampled Block-Sparse Attention.}
We propose a \OUR\ framework that evaluates the block quality in the downsampled attention space, without any finetuning, and improving scalability to ultra-long sequences.

\item \textbf{Norm-Sorting and Covariance Compensation.}
Based on our theoretical analysis, we mitigate the approximation errors via a lightweight norm-sorting module and a covariance-compensated correction, achieving more accurate attention patterns.

\item \textbf{Empirical Validation and Scalability.}
As a result, our \OURshort\ achieves up to 6.95× speedup over FlashAttention and maintains almost full-attention performance at about 50\% sparsity across ultra-long sequence tasks.

\end{itemize}

\section{Preliminaries}
\label{sec:prelim}

\paragraph{Scaled Dot-Product Attention}
As a fundamental component of modern large language models, scaled dot-product attention (SDPA) aggregates contextual information through a weighted summation of value vectors, proving highly effective for modeling long-range dependencies.
Let $Q$ and $K$ be the RoPE-enhanced queries and keys, and $V$ be the values, the attention mechanism is computed as

\begin{equation}
\begin{aligned}
  & A = \mathrm{softmax}\!\left(\frac{QK^\top}{\sqrt{d}}\right), 
\label{eq:sdpa}
\end{aligned}
\end{equation}
where $A_{ij}$ denotes the attention weight from query position $i$ to key position.
This formulation allows queries directly attend to every KV pair, enabling long-range interactions at the cost of quadratic complexity $O(L^2)$.

\paragraph{Block-Sparse Attention}
To reduce the $O(L^2)$ overhead, Flash-Attention~\cite{dao2023flashattention2} reorganizes attention into block-wise computations that leverage fast on-chip memory. 
Let the block size be $B$, and the numbers of query and key blocks as $N_q = \frac{L_q}{B}, N_k = \frac{L_k}{B}$, where $L_q$ and $L_k$ are length of query and key. 
Building on this design, subsequent works further improve the computational efficiency by introducing block-sparse patterns that prune interactions at the block level.

In block-sparse attention, we only compute logits of selected blocks. The block-level binary mask is denoted as $M\in \{0,1\}^{N_q \times N_k}$. For convenience, we use lower-case $i,j$ for token indices, and $g_q,g_k$ for block indices. An item in the block mask $M$ indicates a block pair $(g_q,g_k)$ in the obtained sparse attention map. 
For each query block $Q_{g_q}$, the computation for the corresponding output $O_{g_q}$ is associated with a few number of key-value blocks (\ie $K_{g_k}$,$V_{g_k}$), which are selected by $M_{g_q,g_k}=1$. Thus, specifying a block-sparse attention pattern is to construct a block mask $M$ over $N_q \times N_k$. 

\begin{figure*}[ht] 
  \centering
  \includegraphics[width=1.\linewidth]{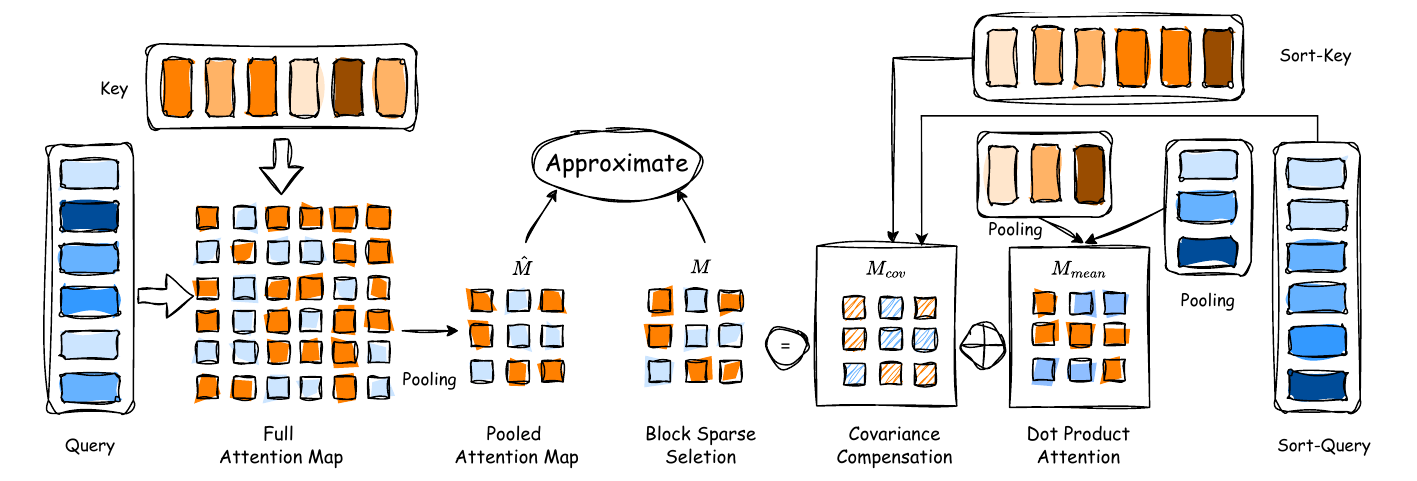} 
  \caption{Illustration of our \OUR. Context length $L=6$ and block size $B=2$ for simplification in this example. The left part shows the post-downsampled oracle block distribution. The right region details the key steps of our \OURshort, including norm-based QK sort, block-wise pooling, and covariance compensation. }
  \label{fig:main}
  \vspace{-3mm}
\end{figure*}

\section{Method}
\label{sec:method}
As is shown in~\cref{fig:main}, we introduce a training-free \emph{block-sparse attention framework} including three core components: a downsampled block-sparse formulation together with an oracle block mask, a norm-based ranking that reshapes blocks to be more homogeneous, and a covariance-based compensation that recovers performance loss.

\subsection{Pre-downsampled Attention Map}
\label{subsec:downsample_oracle}

We construct a downsampled block-sparse baseline to establish a reference formulation.
First, we employ mean pooling to obtain representative block embeddings $\bar{Q}_{g_q}$ and $\bar{K}_{g_k}$, where $g_q,g_k$ are indices of the query and key blocks.
Based on the downsampled representations, we compute block-level logits $\ell_{g_q,g_k}$ and the corresponding query-block distribution $m_{g_q,g_k}$ as :
\begin{equation}
\begin{aligned}
    & \ell_{g_q,g_k}
      = (\bar{Q}_{g_q} \cdot \bar{K}_{g_k}) / \sqrt{d},
       \\
    & m_{g_q,g_k}
    =\mathrm{softmax}(\ell)
      = \exp(\ell_{g_q,g_k}) / \sum_{g_k'=1}^{N_k} \exp(\ell_{g_q,g_k'}),
\label{eq:block-logit}
\end{aligned}
\end{equation}
where $d$ denotes the feature dimension.
The block-sparse attention mask $M$ is subsequently constructed according to the block distribution $m_{g_q,g_k}$ under different computational budgets.
Attention computation is then confined to the selected block pairs, where we employ efficient FlashAttention-style Triton~\cite{triton} kernels to perform the final weighted aggregation.

\vspace{-3mm}
\paragraph{Oracle block distribution.}
To provide an upper bound for analysis and approximation, we define an oracle block distribution that directly aggregates the token-level attention scores in the high-resolution space.
Let $A_{ij}$ denote the attention weight from query token $i$ to key token $j$ within the full attention matrix $A$ from ~\cref{eq:sdpa}. The oracle block mass for block $(g_q,g_k)$ is
\begin{align}
    & \hat{m}_{g_q,g_k}
      = \frac{1}{|I(g_q)|}
        \sum_{i \in I(g_q)}
        \sum_{j \in J(g_k)} A_{ij},
      \qquad
      \label{eq:oracle-dist}
\end{align}
where $I(g_q), J(g_k)$ denotes the index sets belonging to the block. 
Since computing $\hat{m}$ requires the full dense attention map, we use it as the target and an empirical reference for evaluating the fidelity of our approximate block distributions.


\subsection{Downsampling Deviation Analysis}
\label{subsec:error}

Our objective is to make the downsampled attention distribution $m$ as close as the oracle distribution $\hat{m}$.
However, directly analyzing their divergence in the softmax space is analytically intractable due to the normalization coupling and exponential nonlinearity in softmax.
Following prior analyses of approximation errors in attention mechanisms~\cite{choromanski2022rethinkingattentionperformers,dao2022flashattentionfastmemoryefficientexact}, 
we instead examine the deviation at the \emph{pre-softmax logit} level as a surrogate.

The softmax operator is known to be \emph{Lipschitz continuous} under the $\ell_\infty$ norm and smooth under $\ell_2$ perturbations~\cite{boyd2004convex}.
Consequently, a small perturbation in the logits yields a bounded change in the resulting attention distribution.
Formally, for any two logit matrices $\ell$ and $\hat{\ell}$, we have
\begin{equation}
    \| \mathrm{softmax}(\ell) - \mathrm{softmax}(\hat{\ell}) \|_1 
    \le C \, \|\ell - \hat{\ell}\|_\infty,
    \label{eq:softmax-lip}
\end{equation}
where $C$ is a bounded constant depending on the temperature.
Let $\hat{\ell}_{ij}=({Q}_{i} \cdot {K}_{j}) / \sqrt{d}$ represents the pre-softmax token-level logit.
This property implies that controlling the logit-level deviation 
$\big|\hat{\ell}_{ij}-\ell_{g_q,g_k}\big|$ 
provides a valid proxy for bounding the difference between 
$m$ and $\hat{m}$.
Therefore, we focus our analysis on the geometric factors that govern this logit deviation, which further motivates the proposed norm-sorting and covariance-compensated designs.

We rewrite $Q_i=\bar Q_{g_q}+\delta Q_i$, $K_j=\bar K_{g_k}+\delta K_j$. Then, the deviation between the token-level logit $\hat{\ell}_{ij}$ and its corresponding block-level logit $\ell_{g_q,g_k}$ in~\cref{eq:block-logit}, is written as:

\begin{equation}
    \hat{\ell}_{ij}-\ell_{g_q,g_k}
    =\frac{1}{\sqrt d}\big(\delta Q_i\!\cdot\!\bar K_{g_k}
    +\bar Q_{g_q}\!\cdot\!\delta K_j+\delta Q_i\!\cdot\!\delta K_j\big).
    \label{eq:logit-deviation}
\end{equation}

By applying Cauchy--Schwarz, we obtain a formulation as :
\begin{equation}
\begin{aligned}
& |\delta Q_i\!\cdot\!\bar K_{g_k}|\le R^Q_{g_q}M^K_{g_k}, \\
& |\bar Q_{g_q}\!\cdot\!\delta K_j|\le M^Q_{g_q}R^K_{g_k}, \\
& |\delta Q_i\!\cdot\!\delta K_j|\le R^Q_{g_q}R^K_{g_k}.
\end{aligned}
\end{equation} 
In this way, we define several block-wise norm metrics as :
\begin{equation}
\begin{aligned}
    & R^Q_{g_q}
      = \max_{i \in I(g_q)} \big\|Q_i - \bar{Q}_{g_q}\big\|_2, 
    \qquad
     M^Q_{g_q}
      = \max_{i \in I(g_q)} \|Q_i\|_2, \\
    & R^K_{g_k}
      = \max_{j \in J(g_k)} \big\|K_j - \bar{K}_{g_k}\big\|_2, 
      \qquad
    M^K_{g_k}
      = \max_{j \in J(g_k)} \|K_j\|_2,
\end{aligned}
\end{equation}
where $R^Q_{g_q}$ and $R^K_{g_k}$ are \emph{radii} measuring how far tokens deviate from their block means
$\bar Q_{g_q}$ and $\bar K_{g_k}$, and $M^Q_{g_q}$ and $M^K_{g_k}$ are \emph{max norms}, which capture the scale inside blocks.
Then, max norms are combined with upper bounds on the deviation between token-level and block-level logits $U_{g_q,g_k}$,
\begin{equation}
\begin{aligned}
    &U_{g_q,g_k}=\frac{R^Q_{g_q}M^K_{g_k}+M^Q_{g_q}R^K_{g_k}+R^Q_{g_q}R^K_{g_k}}{\sqrt d},\\
    &\big|\hat{\ell}_{ij}-\ell_{g_q,g_k}\big|
    \le U_{g_q,g_k}\qquad(\forall\, i\in I(g_q),\, j\in J(g_k)).
    \end{aligned}
    \label{eq:logits-bound}
\end{equation}

\begin{figure}[t]
  \centering
  \begin{subfigure}{0.47\linewidth}
    \centering
    \includegraphics[width=\linewidth]{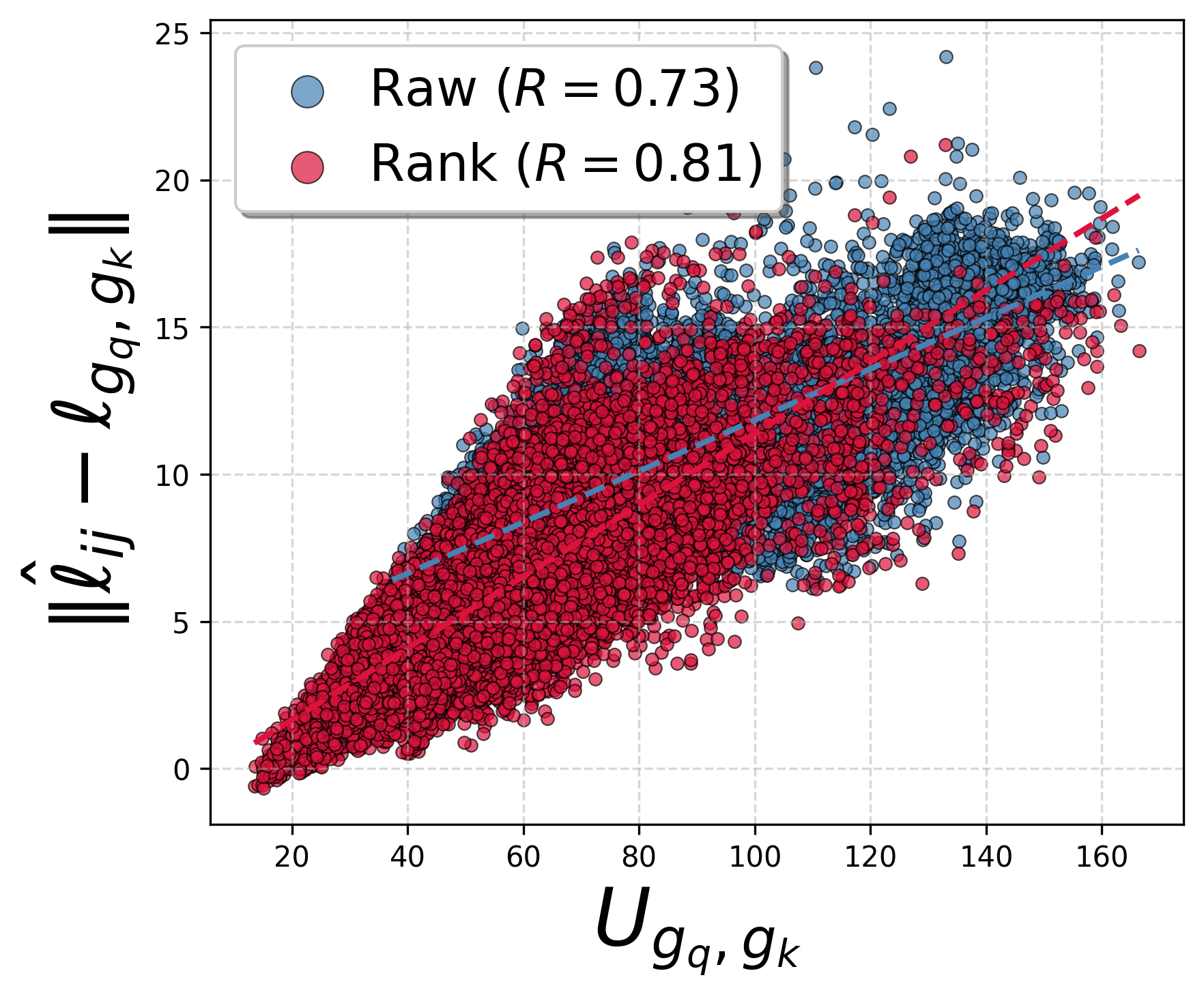}
    \caption{LLaDA-V on VideoMME.}
  \end{subfigure}
  \begin{subfigure}{0.47\linewidth}
    \centering
    \includegraphics[width=\linewidth]{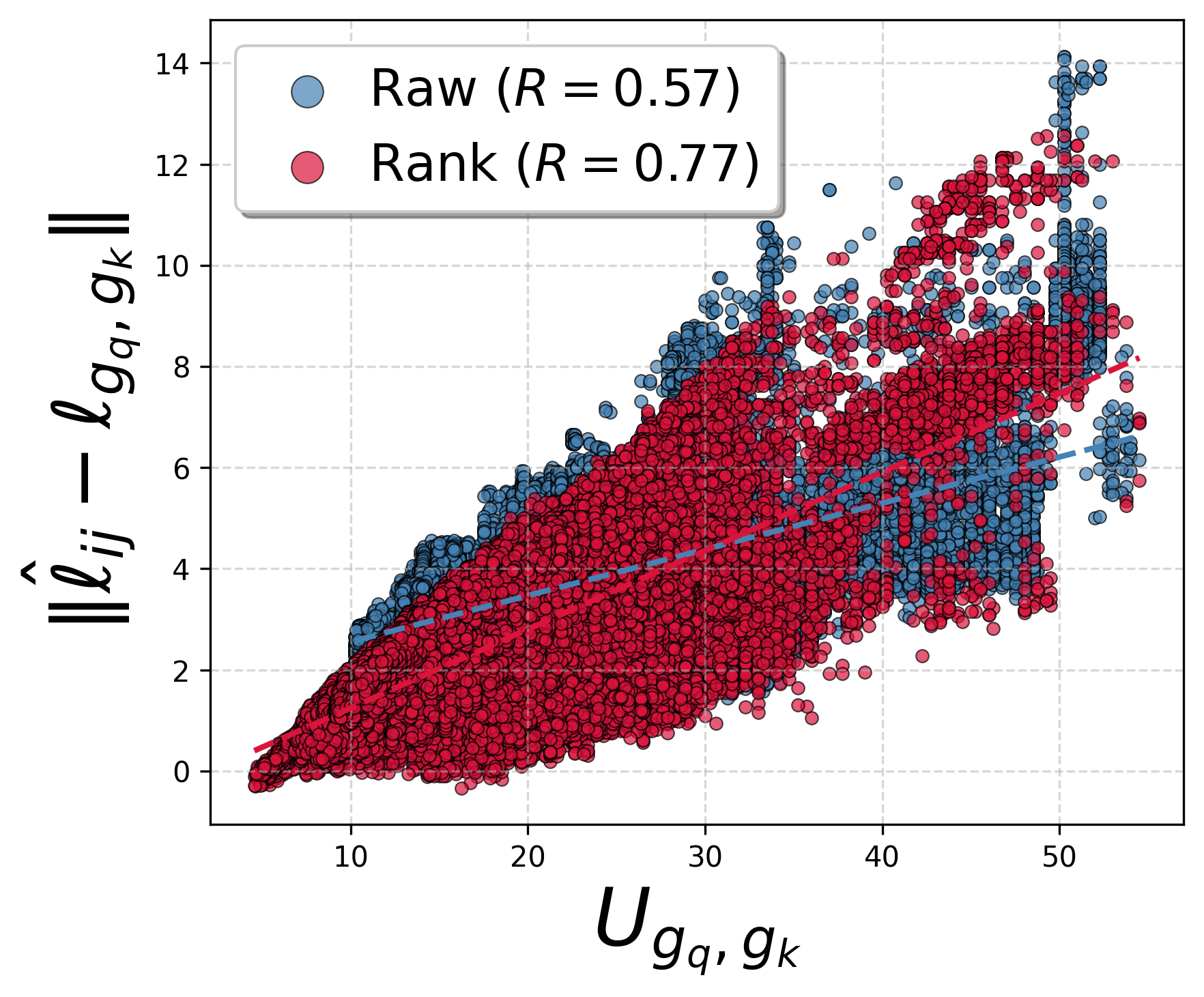}
    \caption{UltraLLaDA on Ruler.}
  \end{subfigure}
  \caption{Correlation between the Norm-based Metric $U_{g_q,g_k}$ and logit deviation $\big|\hat{\ell}_{ij}-\ell_{g_q,g_k}\big|$ across the layers and heads. The blue group represents the vanilla data distribution, and the red group indicates the data after norm-based sorting. The correlation coefficient $R$ of each group shows the linear association.}
  \label{fig:corr}
  \vspace{-3mm}
\end{figure}

\vspace{-6mm}
\paragraph{Interpretation and visualization.}
In principle, a smaller radii ($R^Q,R^K$) and moderate max norms ($M^Q,M^K$) would make blocks homogeneous, thus improving performance.
If $R^Q_{g_q}, R^K_{g_k}$ are zeros, standing for perfectly homogeneous blocks, $U_{g_q,g_k}$ would be zero and token-level logits match block logits. 
As shown in~\cref{fig:corr}, we visualize the relationship between the theoretical upper bound $U_{g_q,g_k}$ and the observed maximum logit deviation $\max_{i,j} |\hat{\ell}_{ij} - \ell_{g_q,g_k}|$.  $R \in [-1, 1]$ is the Pearson correlation coefficient produced by the linear regression model. Blue group represents vanilla data distribution, and $R>0.5$ indicates that $U_{g_q,g_k}$ is an informative proxy for the true approximation deviation across blocks.

\subsection{Norm-Based Ranking}
\label{subsec:norm-rank}
The previous analysis establishes that the deviation between the oracle and pre-downsampled attention distributions 
is governed by the upper bound $U_{g_q,g_k}$ in~\cref{eq:logits-bound}, 
which depends on the intra-block \emph{radii} $(R^Q, R^K)$ and \emph{max norms} $(M^Q, M^K)$. 
Intuitively, when tokens inside a block exhibit highly uneven norm magnitudes, 
the resulting query--key interactions become heterogeneous, 
leading to unstable logits and dispersed attention patterns. 
Empirical observations in~\cref{fig:corr} confirm that blocks with large $U_{g_q,g_k}$ 
correlate strongly with greater logit deviations and degraded approximation fidelity.

However, under the \emph{training-free} constraint, 
we cannot rely on learnable gates or additional supervision (as in SeerAttention~\cite{seeratt}) 
to calibrate these heterogeneous regions. 
Therefore, a natural question arises: 
\emph{Can we reshape the block composition itself to reduce $U_{g_q,g_k}$ 
and make the attention map more homogeneous, without introducing any trainable parameters?} 

To this end, we propose a lightweight \textbf{norm-based ranking} strategy 
that reorders tokens by their feature norm before block partitioning.
This operation effectively clusters high-activation tokens together, 
reducing the intra-block variance of $QK$ magnitudes and tightening the theoretical bound $U_{g_q,g_k}$. 
Given a key sequence, each key token $j \in \{1,\dots,L_k\}$, define a scalar score,
\begin{align}
    & s_j^K = \|K_j\|_2.
\end{align}
Let $\pi_k$ be a permutation that sorts token norms in non-decreasing order
\begin{align}
    & s^K_{\pi_k(1)} \le s^K_{\pi_k(2)} \le \dots \le s^K_{\pi_k(L_k)},
\end{align}
and define the permuted query sequence $K'_i = K_{\pi_k(i)}$.
Then, we form new query blocks on $K'$ by grouping consecutive indices of size $B$.
We can analogously permute queries using $s_i^q = \|Q_i\|_2$ to obtain $Q'$.
Under common heavy-tailed activation patterns, sorting by norm and grouping noticeably decreases within-block variance and radii.

\vspace{-3mm}
\paragraph{Effect on the deviation bound.}
In practice, norm-based ranking substantially reduces the empirical distribution of $R^{Q}_{g_q}$ and $R^{K}_{g_k}$, thereby shrinking the theoretical upper bound $U_{g_q,g_k}$ and narrowing the gap between block-wise attention estimation and the oracle full-attention baseline.
We visualize the effect by comparing the relationship between the theoretical bound $U_{g_q,g_k}$ and the actual maximum logit deviation $\|\hat{\ell}_{ij} - \ell_{g_q,g_k}\|$ before (blue) and after (red) applying norm-based ranking in Fig.~\ref{fig:corr}. 
The improvement of the correlation coefficient $R$ shows that norm-based ranking not only reduces the magnitude of the bound but also enhances its predictive power, making it a more faithful proxy for the true approximation error.
Importantly, this process is entirely deterministic and parameter-free, 
aligning with our training-free design philosophy while providing a principled means 
to control the approximation error between $m$ and $\hat{m}$.


\subsection{Covariance-Based Compensation}
\label{subsec:cov-comp}

\paragraph{Recalling the logit decomposition.}
From the token/block logit relation in ~\cref{eq:logit-deviation}, we have
\begin{equation}
\hat{\ell}_{ij}-\ell_{g_q,g_k}
=\frac{1}{\sqrt d}\Big(
\underbrace{\delta Q_i\!\cdot\!\bar K_{g_k}}_{\text{(I)}}
+\underbrace{\bar Q_{g_q}\!\cdot\!\delta K_j}_{\text{(II)}}
+\underbrace{\delta Q_i\!\cdot\!\delta K_j}_{\text{(III)}}
\Big).
\label{eq:logit-decomp}
\end{equation}
By construction $\mathbb{E}_{i\in I(g_q)}[\delta Q_i]=\mathbf{0}$ and $\mathbb{E}_{j\in J(g_k)}[\delta K_j]=\mathbf{0}$,
so the \emph{block expectation} of (I) and (II) vanishes, whereas (III) may leave a \emph{systematic bias}:
\begin{equation}
\mathbb{E}_{i,j}\!\big[\hat{\ell}_{ij}-\ell_{g_q,g_k}\big]
=\frac{1}{\sqrt d}\,\mathbb{E}_{i,j}\!\big[\delta Q_i^\top\delta K_j\big]
\ \approx\ \frac{1}{d}\,\mathrm{tr}\!\big(\Sigma^Q_{g_q}\Sigma^K_{g_k}\big),
\label{eq:cov-bias}
\end{equation}
where $\Sigma^Q_{g_q}=\mathbb{E}_i[\delta Q_i\delta Q_i^\top]$ and
$\Sigma^K_{g_k}=\mathbb{E}_j[\delta K_j\delta K_j^\top]$.
Equation~\eqref{eq:cov-bias} reveals that even after reducing the variance-like terms via norm-based ranking
(\ie tightening $U_{g_q,g_k}$ in \cref{eq:logits-bound}),
a \emph{second-order} covariance contribution can still bias the block logits.

\vspace{-3mm}
\paragraph{Designing the compensation term.}
Guided by \eqref{eq:cov-bias}, we add a lightweight second-order correction to the block logit:
\begin{equation}
\tilde{\ell}_{g_q,g_k}
=\ell_{g_q,g_k}\;+\;\beta\,\Delta_{g_q,g_k},
\Delta_{g_q,g_k}
:=\frac{1}{d}\,\mathrm{tr}\!\big(\Sigma^Q_{g_q}\Sigma^K_{g_k}\big),
\label{eq:cov-comp}
\end{equation}
where $\beta$ is a scalar to balance fidelity and stability (default $\beta{=}1$).
Intuitively, $\Delta_{g_q,g_k}$ upweights blocks whose query/key residuals are more correlated,
recovering contrast that is lost when using block means alone.

\vspace{-3mm}
\paragraph{Diagonal (variance) approximation and complexity.}
Computing $\mathrm{tr}(\Sigma^Q\Sigma^K)$ exactly costs $O(Ld^2)$.
In practice, we adopt a diagonal-variance approximation that yields three additive terms:
\begin{equation}
\begin{aligned}
\Delta_{g_q,g_k} \approx&  \frac{1}{d} \sum_{t=1}^{d} \Big( 
\mathrm{Var}[Q_t]_{g_q} \, \bar{K}_{g_k,t}^2\\
+& \mathrm{Var}[K_t]_{g_k} \, \bar{Q}_{g_q,t}^2
+ \mathrm{Var}[Q_t]_{g_q} \, \mathrm{Var}[K_t]_{g_k}
\Big).
\end{aligned}
\label{eq:diag-variance-form}
\end{equation}
It only requires blockwise first/second moments (computed alongside $\bar Q,\bar K$),  adds $O((N_q{+}N_k)d)$ extra work per layer, and keeps the overall complexity $O(Ld)$.

\vspace{-6mm}
\paragraph{Summary.}
Norm-based ranking primarily \emph{reduces variance} (tightening $U_{g_q,g_k}$),
while covariance-based compensation \emph{removes residual bias} predicted by \eqref{eq:logit-decomp}.
The two are complementary: together they narrow the gap between the practical downsampled distribution $m$
and the oracle $\hat m$ without any training or dense-map construction.
Finally, we summarize the \OUR\ framework and provide pseudo-code in~\cref{alg:nrcc}.

\begin{algorithm}[t]
\caption{Norm-Rank Attention with Covariance Compensation (\OUR)}
\label{alg:nrcc}
\begin{algorithmic}[1]
\REQUIRE Inputs ${Q}, {K}$, $V$; 
         block size $B$; per-query-block budget $\kappa$; 
         compensation weight $\beta$.
\vspace{2pt}
\STATE \textbf{Query Ranking by Norm:}
       compute $s_i^Q \!=\! \|Q_i\|_2$, obtain permutation $\pi_q$ by (windowed) sort, 
       and reorder $Q'_i \!\leftarrow\! Q_{\pi_q(i)}$
\STATE \textbf{ Key Ranking by Norm:}
       compute $s_j^K \!=\! \|K_j\|_2$, obtain $\pi_k$, 
       and reorder $K'_j \!\leftarrow\! K_{\pi_k(j)}$, $V'_j \!\leftarrow\! V_{\pi_k(j)}$
\STATE \textbf{Block Partition:}
       divide $Q', K'$ into blocks 
       $\{I(g_q)\}_{g_q=1}^{N_q}$ and $\{J(g_k)\}_{g_k=1}^{N_k}$ of size $B$
\STATE \textbf{Block Statistics:}
       for each $(g_q, g_k)$, compute block means 
       $\bar{Q}_{g_q}, \bar{K}_{g_k}$ and covariances 
       $\Sigma^Q_{g_q}, \Sigma^K_{g_k}$
\STATE \textbf{Baseline Logits:}
       $\ell_{g_q,g_k} = \frac{1}{\sqrt{d}}\, 
        \bar{Q}_{g_q} \!\cdot\! \bar{K}_{g_k}$
\STATE \textbf{Covariance Compensation:}
       compute $\widehat{\Delta}_{g_q,g_k}$ in~\cref{eq:diag-variance-form}
\STATE \textbf{Compensated Block Logits:}
       $\ell'_{g_q,g_k}
        \leftarrow \ell_{g_q,g_k} 
        + \beta \,\widehat{\Delta}_{g_q,g_k}$
\FOR{each query block $g_q = 1, \dots, N_q$}
    \STATE $m'_{g_q,\cdot}
           \leftarrow \mathrm{softmax}(\{\ell'_{g_q,g_k}\}_{g_k})$
    \STATE select top-$\kappa$ key blocks by $m'_{g_q,\cdot}$; 
           set $M_{g_q,g_k} = 1$ if selected, else $0$
\ENDFOR
\STATE \textbf{Block Sparse Attention Computation:}
       execute block-sparse attention on  $Q', K', V'$  over
       $\{(g_q,g_k) : M_{g_q,g_k}=1\}$ using Triton kernels;
       remap outputs to the original order via $\pi_q^{-1}$
\RETURN Sparse attention outputs
\end{algorithmic}
\end{algorithm}

\section{Experiments}
\label{sec:Experiments}

\subsection{Experimental Settings}

\newcommand{\rotangle}{60} %
\newcommand{\rotheader}[1]{\rotatebox{\rotangle}{\scriptsize #1}} %

\newcommand{\rottt}[1]{\rotatebox[origin=c]{75}{#1}}

\begin{table*}[t]
\setlength{\tabcolsep}{2pt}
\centering
\small
\caption{Comparison of different attention methods on the LongBench benchmark truncated to a 16K context length, implemented with LLaDA1.5 and UltraLLaDA. AVG is an aggregated question-count–weighted average score.}
\vspace{-2mm}
\label{tab:longbench}
\begin{tabular}{l cccc cccc cccc cccc ccc cc c}
\toprule
 & \multicolumn{4}{c}{Single-Doc QA} & \multicolumn{4}{c}{Multi-Doc QA} & \multicolumn{4}{c}{Summarization} & \multicolumn{4}{c}{Few-shot Learning}& \multicolumn{3}{c}{Synthetic}  & \multicolumn{2}{c}{Code} &  \\
\cmidrule(lr){2-5} \cmidrule(lr){6-9} \cmidrule(lr){10-13} \cmidrule(lr){14-17} \cmidrule(lr){18-20}  \cmidrule(lr){21-22}
Method & \rottt{NrtvQA} & \rottt{Qasper} & \rottt{MF-en} & \rottt{MF-zh}& 
\rottt{HPQA} & \rottt{2WikiMQA} & \rottt{MuSiQue}& \rottt{DuReader} & 
\rottt{GovReport} & \rottt{QMSum} & \rottt{MultiNews} & \rottt{VCSum} & 
\rottt{TREC} & \rottt{TriviaQA} & \rottt{SAMSum} & \rottt{LSHT} & 
\rottt{PassCount} & \rottt{PassRe-en} & \rottt{PassRe-zh} & 
\rottt{LCC} & \rottt{RB-P} & \textbf{AVG} \\
\midrule
LLaDA1.5 & 4.1 & \textbf{20.4} & \textbf{21.8} & \textbf{20.8} & 6.3 & \textbf{8.1} & 3.4 & 10.5 & 17.3 & 5.4 & 14.3 & 6.7 & \textbf{73.5} & \textbf{63.6} & 32.2 & 21.5 & 1.9 & \textbf{29.5} & \textbf{70.9} & \textbf{68.0} & {60.6} & \textbf{31.5} \\
+XAtt & 4.3 & 20.3 & 21.5 & 19.9 & \textbf{7.5} & 7.9 & 3.5 & 10.8 & 16.8 & \textbf{5.8} & 14.2 & \textbf{7.5} & 61.5 & 61.5 & 28.7 & 19.3 & 1.9 & 27.7 & 56.8 & 61.9 & 55.0 & 28.8 \\
\rowcolor{cyan!15}
+Ours & \textbf{4.9} & \textbf{20.4} & \textbf{21.8} & 20.1 & 6.3 & 8.0 & \textbf{3.7} & \textbf{11.2} & \textbf{17.4} & 5.4 & \textbf{14.4} & 6.7 & 71.0 & 60.1 & \textbf{33.6} & \textbf{22.3} & \textbf{2.1} & 28.4 & 68.2 & 67.9 & \textbf{60.8} & 31.3 \\
\midrule
UltraLLaDA & {12.1} & \textbf{14.5} & 27.4 & 19.9 & 13.4 & \textbf{11.0} & \textbf{9.7} & \textbf{20.4} & \textbf{20.8} & 7.8 & 14.7 & 7.5 & \textbf{79.5} & \textbf{92.0} & \textbf{36.2} & \textbf{40.5} & 0.7 & 81.3 & \textbf{72.9} & 66.2 & 56.9 & \textbf{37.2} \\
+XAtt & 4.1 & 13.9 & 25.2 & {20.1} & 6.2 & 6.8 & 5.2 & 12.9 & 19.9 & 7.0 & \textbf{16.0} & 7.4 & 75.0 & 85.2 & 33.3 & 28.5 & \textbf{2.1} & 81.4 & 64.6 & 64.5 & \textbf{63.5} & 34.9 \\
\rowcolor{cyan!15}
+Ours & \textbf{13.3} & 13.6 & \textbf{27.9} & \textbf{20.3} & \textbf{13.8} & 9.8 & 8.7 & 12.6 & \textbf{20.8} & \textbf{7.9} & 12.7 & \textbf{8.3} & 75.5 & 90.9 & 35.5 & 34.4 & 1.3 & \textbf{82.3} & 70.9 & \textbf{69.0} & 63.1 & \textbf{37.2} \\
\bottomrule
\end{tabular}
\end{table*}

\begin{table*}[t]
\setlength{\tabcolsep}{9.8pt}
\centering
\small
\caption{
Comparison of different attention methods on the RULER benchmark across context lengths from 4K to 32K, using LLaDA1.5 and UltraLLaDA as base models. Evaluated categories include: Retrieval (Needle-in-a-Haystack, NIAH), Aggregation (frequent word extraction, AGG), Question Answering (QA), and Multi-hop Tracing (variable tracking, VT). AVG denotes a question-count-weighted average score across all categories. ``--'' indicates failure.}
\vspace{-2mm}
\label{tab:ruler_results}

\begin{tabular}{l | c c c c |c | c c c c |c}
\toprule
\textbf{Model} & \multicolumn{5}{c |}{\textbf{4K}} & \multicolumn{5}{c}{\textbf{8K}} \\
\hline 
 & {NIAH} & {AGG} & {QA} & {VT} & \textbf{AVG}  & {NIAH} & {AGG} & {QA} & {VT} & \textbf{AVG}\\
\hline
LLaDA1.5      & \textbf{99.25} &\textbf{47.78} & \textbf{88.5} & \textbf{100} & \textbf{89.74}    &\textbf{ 51.56} & 41.6 & 48.02 &\textbf{61.6} &50.25 \\
LLaDA1.5+XAtt & 74.66 & 26.55 & 61.00 &3.00 & 59.64  &39.53 & 33.21 & 35.5 & 0.4 &34.93 \\
\rowcolor{cyan!15}
LLaDA1.5+Ours & 98.16 & 41.20 & 85.5 & 99.4 & 87.54  & 50.28 & \textbf{58.23} & \textbf{48.1} & 59.4 & \textbf{51.85} \\
\hline
UltraLLaDA & 97.31 & 66.69 & 68.50 & \textbf{100} & 88.37    & 98.28 & 55.29 & 62.00 & \textbf{100} & 86.22 \\
UltraLLaDA+XAtt &95.12 &49.61 & 70.50 &57.40 &81.43  & 86.97 &36.30 & 62.00 & 24.20 &70.50 \\
\rowcolor{cyan!15}
UltraLLaDA+Ours &\textbf{98.16} &\textbf{67.78} & \textbf{75.00}  & \textbf{100} &\textbf{90.06}  & \textbf{98.5} & \textbf{57.65} & \textbf{68.50} & \textbf{100} & \textbf{87.71} \\

\midrule \midrule 

\textbf{Model} & \multicolumn{5}{c |}{\textbf{16K}} & \multicolumn{5}{c}{\textbf{32K}} \\
\hline 
 & {NIAH} & {AGG} & {QA} & {VT} & \textbf{AVG}  & {NIAH} & {AGG} & {QA} & {VT} & \textbf{AVG} \\
\hline
LLaDA1.5 &  8.56 &11.19 & 46.5 & 20.6 &15.73 &-- &--&--&--&-- \\
LLaDA1.5+XAtt &15.16 & 15.89 & 37.00 & 0.6 &17.51 &-- &--&--&--&--  \\
\rowcolor{cyan!15}
LLaDA1.5+Ours &\textbf{16.5} & \textbf{18.06} & \textbf{74.00} & \textbf{30.00} &\textbf{26.62} &-- &--&--&--&--\\
\hline
UltraLLaDA & 93.00 & 43.12 & 39.50 & 98.40 &77.51    & \textbf{92.78} & \textbf{29.29} & 29.00 & 98.40 & \textbf{73.63} \\
UltraLLaDA+XAtt  &75.09&35.37&40.00 &7.40 &58.38   &74.94 &30.19 &15.50 &1.80&53.28 \\
\rowcolor{cyan!15}
UltraLLaDA+Ours  &\textbf{94.62}&\textbf{44.89}&\textbf{53.00}&\textbf{99.40} &\textbf{80.93}   &90.97 &27.85 &\textbf{32.00} &\textbf{100}&72.88 \\
\bottomrule
\end{tabular}
\vspace{-5mm}
\end{table*}

\vspace{-2mm}
\paragraph{Baselines and Platforms}
We evaluate our block-sparse attention framework across three distinct domains. (1) For natural language tasks, we employ LLaDA1.5~\cite{llada1.5} and UltraLLaDA~\cite{ultrallada}. (2) In the multi-modal understanding domain, we utilize LLaDA-V~\cite{lladav}. (3) Finally, for video generation, we use the Wan2.1-T2V-14B~\cite{wan2.1}. 
Our dense attention baseline is implemented using \textsc{FlashAttention-2}~\cite{dao2023flashattention2} within the \textsc{FlashInfer}~\cite{flashinfer} framework. We utilize \textsc{SVG2}~\cite{svg2} as a strong baseline for video generation.
All experiments are conducted on the NVIDIA A100 (80GB) GPU platform using PyTorch.
We optimize kernel performance via Triton with a block size of 128 and use greedy decoding.  
For \textsc{UltraLLaDA}, \textsc{LLaDA1.5}, and \textsc{LLaDA-V}, we set the sparsity over 50\% and integrate the \textsc{Fast-DLLM}~\cite{wu2025fastdllmtrainingfreeaccelerationdiffusion} for fast evaluation.

\vspace{-3mm}
\paragraph{Datasets and Implementation}
We evaluate our model across three domains: natural language understanding, vision understanding, and video generation.
For Natural Language Understanding, we use the RULER dataset \cite{hsieh2024ruler} to test long-context reasoning with controllable sequence lengths and task complexities, and the LongBench benchmark \cite{bai2023longbench} for real-world long-context evaluation. Second, we adopt ChartQA \cite{masry2022chartqa}, DocVQA \cite{mathew2021docvqa}, InfoVQA \cite{mathew2022infographicvqa}, RealworldQA \cite{x2024realworldqa}, MMMU \cite{yue2024mmmu}, and MMMU-Pro \cite{yue2024mmmupro} for image–language tasks.
Next, for multi-image and video understanding, we evaluate on Video-MME \cite{fu2024video}, MLVU-dev \cite{zhou2024mlvu}, and MuirBench \cite{wang2024muirbench}, which stress-test sparse attention under long visual contexts. Notably, Video-MME includes 900 videos (totaling 254 hours) and serves as a large-scale benchmark for multi-modal video comprehension.
Finally, in video generation tasks, we use 96 text prompts from VBench \cite{huang2023vbench} to generate videos, comparing with a full-attention baseline to assess generation.

\subsection{Performance Evaluation}

\begin{figure*}[ht] 
  \centering
  \includegraphics[width=.98\linewidth]{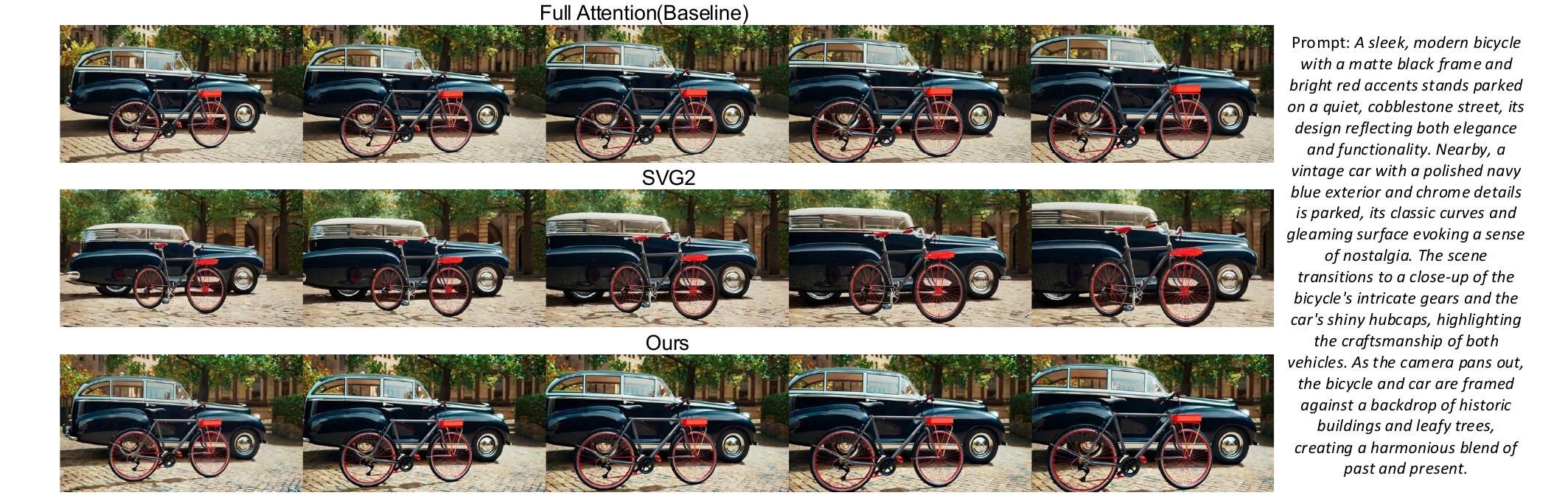} 
  \caption{Qualitative comparison of video generation results on the VBench benchmark dataset. Rows show frames from videos generated using: (1) Full Attention (baseline), (2) SVG2, (3) Ours. Our method achieves high visual fidelity to the full attention baseline.}
  \label{fig:wan} 
  \vspace{-3mm}
\end{figure*}
\paragraph{Performance on Ultra-Long NLP Tasks.}
Our method effectively improves long-context performance across both LLaDA-1.5 and UltraLLaDA models, as shown in \cref{tab:ruler_results}. 
UltraLLaDA+Ours achieves the best overall results, demonstrating that our attention mechanism effectively preserves task-critical information while enabling efficient computation.
As shown in \cref{tab:longbench}, our method demonstrates strong \textit{task-agnostic} effectiveness: it preserves or slightly enhances performance across all categories without task-specific tuning. This contrasts with XAttention, whose fixed sparsity pattern leads to inconsistent gains and notable drops in NarrativeQA and PassRe-zh tasks. The results suggest that our sorting and compensation scheme generalizes well beyond controlled benchmarks to diverse, practical long-context scenarios.

\vspace{-6mm}
\paragraph{Performance on Multimodal Understanding}
On the video understanding benchmarks (\cref{tab:videomme}), our method achieves consistent gains over the LLaDA-V baseline across all datasets, achieving slightly better results on VideoMME and MLVU-dev.

In \cref{tab:vlm}, we show the result on general image understanding tasks. LLaDA-V+Ours matches or exceeds the strong LLaDA-V baseline across diverse tasks—from chart and document understanding to complex multi-modal QA—while significantly outperforming sparsity-based alternatives like Flex and XAtt. Notably, our approach achieves the highest overall average, demonstrating that our prior-free sparse pattern preserves critical visual-semantic information without sacrificing efficiency.

\begin{table}[t] 
\centering
\small
\caption{Comparison of different methods on LLaDA-V in video understanding tasks. All metrics are reported as percentages. Higher is better. AVG is the average score across all datasets.}
\vspace{-2mm}
\label{tab:videomme} 

\begin{tabular}{l cccc}
\toprule
\textbf{Model} & \textbf{VideoMME}&\textbf{MLVU-dev}  & \textbf{MuirBench}\\
\midrule
LLaDA-V      & 56.07 & 59.61 &\textbf{48.12} \\
LLaDA-V+Flex & 48.22 & 47.91 &34.35 \\
LLaDA-V+XAtt & 55.63 & 53.84 & 46.54\\
\rowcolor{cyan!15}
LLaDA-V+Ours & \textbf{56.56}& \textbf{59.71} &47.69 \\
\bottomrule
\end{tabular}
\vspace{-6mm}
\end{table}

\vspace{-3mm}
\paragraph{Performance on Video Generation}
We conduct experiments on the VBench using the Wan2.1 model, with a 10-step full-attention warmup to ensure stable generation. The generated videos 
are produced using 50 denoising steps, resulting in an input sequence of approximately 75K tokens in context length. As shown in \cref{tab:wan2.1}, our approach achieves significantly higher visual fidelity than SVG2, which is a dedicated video generation accelerator, across all metrics—especially at 60\% and 50\% sparsity. It improves PSNR by over 2.5 dB and reduces LPIPS by 35\%. Qualitatively, as illustrated in \cref{fig:wan}, our method generates videos with consistent object appearance and accurate fine-grained details, closely matching the full-attention baseline while maintaining sparsity and efficiency.

\begin{table}[t]
\setlength{\tabcolsep}{4.2pt}
\centering
\small
\caption{Quantitative results of applying Sparse Attention methods to the Wan2.1 model on the VBench benchmark, using a 10-step full-attention warmup. Higher ($\uparrow$) yields better fidelity at the cost of slightly reduced sparsity (higher density). }
\vspace{-2mm}
\label{tab:wan2.1}

\begin{tabular}{l cccc}
\toprule
\textbf{Methods}  & \textbf{PSNR} ($\uparrow$) & \textbf{SSIM} ($\uparrow$) & \textbf{MS\_SSIM} ($\uparrow$)  & \textbf{LPIPS} ($\downarrow$) \\
\midrule
SVG2         & 21.51     & 0.762      & 0.832     & 0.173                 \\
\rowcolor{cyan!15}
Ours(60\%)   & 22.34     & 0.782      & 0.865     &    0.163            \\
\rowcolor{cyan!15}
Ours(50\%)   & \textbf{24.08}     & \textbf{0.833}     & \textbf{0.906}     & \textcolor{red}{\textbf{ 0.112}}            \\

\bottomrule
\end{tabular}
\vspace{-6mm}
\end{table}

\begin{table*}[t] 
\setlength{\tabcolsep}{7.7pt}
\centering
\small
\caption{Comparison of different methods on LLaDA-V in various visual understanding tasks. All metrics are reported as percentages. Higher is better. AVG is the average score across all datasets.}
\vspace{-2mm}
\label{tab:vlm} 

\begin{tabular}{l ccccccccc}
\toprule
\textbf{Model} & \textbf{ChartQA} & \textbf{DocVQA$_{\text{(val)}}$} & \textbf{InfoVQA$_{\text{(val)}}$} & 
\textbf{RealworldQA} & \textbf{MMMU$_{\text{(val)}}$} & 
\textbf{MMMU$_{\text{(ProVision)}}$} & \textbf{AVG} \\
\midrule
LLaDA-V       & \textbf{77.8}  & \textbf{83.48} & \underline{65.02} & 63.01 & \textbf{48.67} & \textbf{18.60} & \underline{59.43} \\
LLaDA-V + Flex & 54.0 & {30.29} & 26.75 & 48.76 & 44.91 & 14.51 & 36.54 \\
LLaDA-V + XAtt & 72.56 & 82.31 & 62.88 & \underline{63.14} & 44.44 & 16.18 & 56.92 \\
\rowcolor{cyan!15}
LLaDA-V + Ours & \underline{77.72} & \underline{82.89} & \textbf{65.26} & \textbf{64.84} & \underline{48.32} & \underline{17.92} & \textbf{59.49}\\
\bottomrule
\end{tabular}
\vspace{-3mm}
\end{table*}

\subsection{Ablation Study}
\begin{figure}[t]
  \centering
  \begin{subfigure}{0.49\linewidth}
    \centering
    \includegraphics[width=\linewidth]{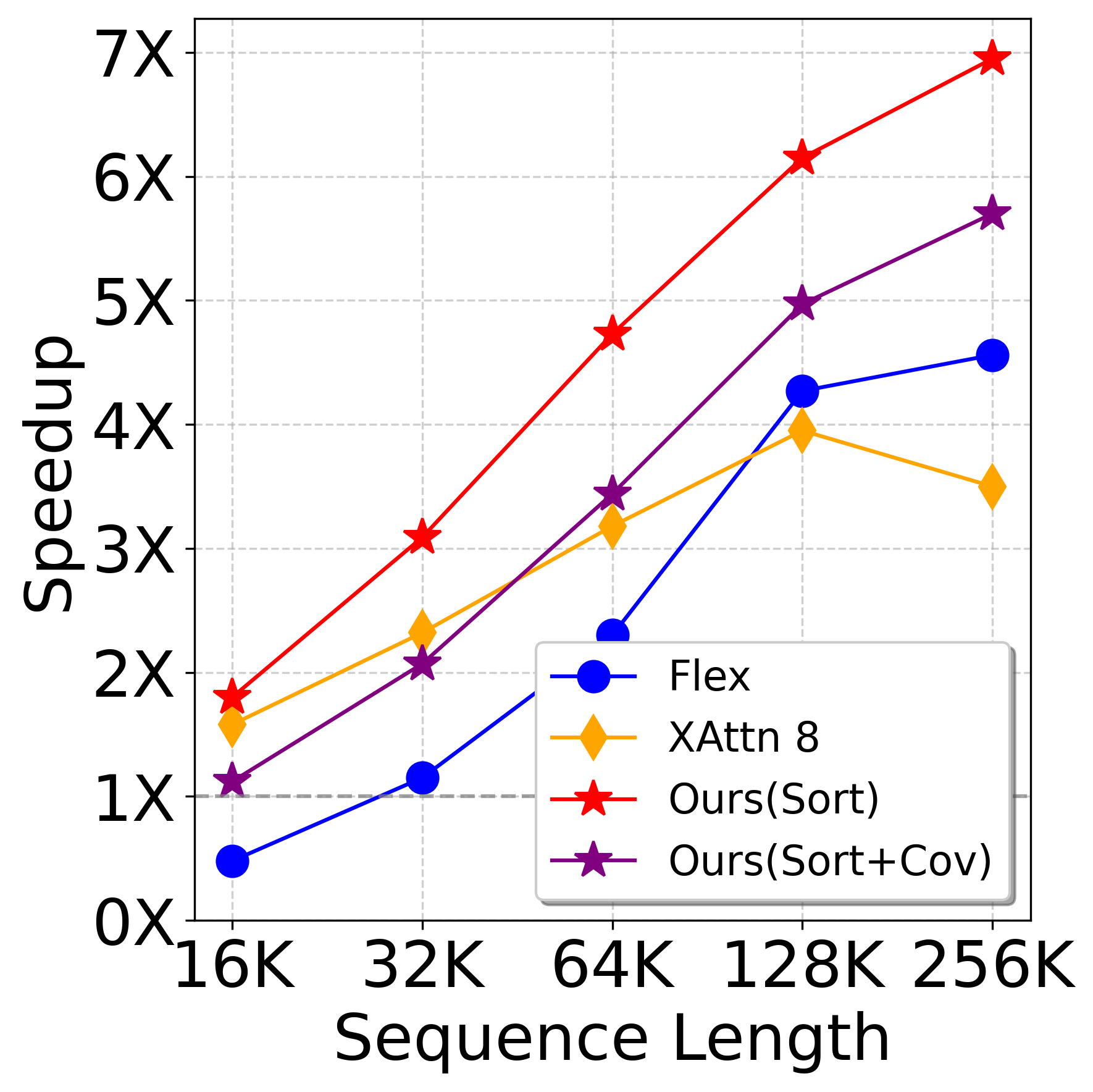}
    \caption{Speedup on A100.}
  \end{subfigure}
  \begin{subfigure}{0.49\linewidth}
    \centering
    \includegraphics[width=\linewidth]{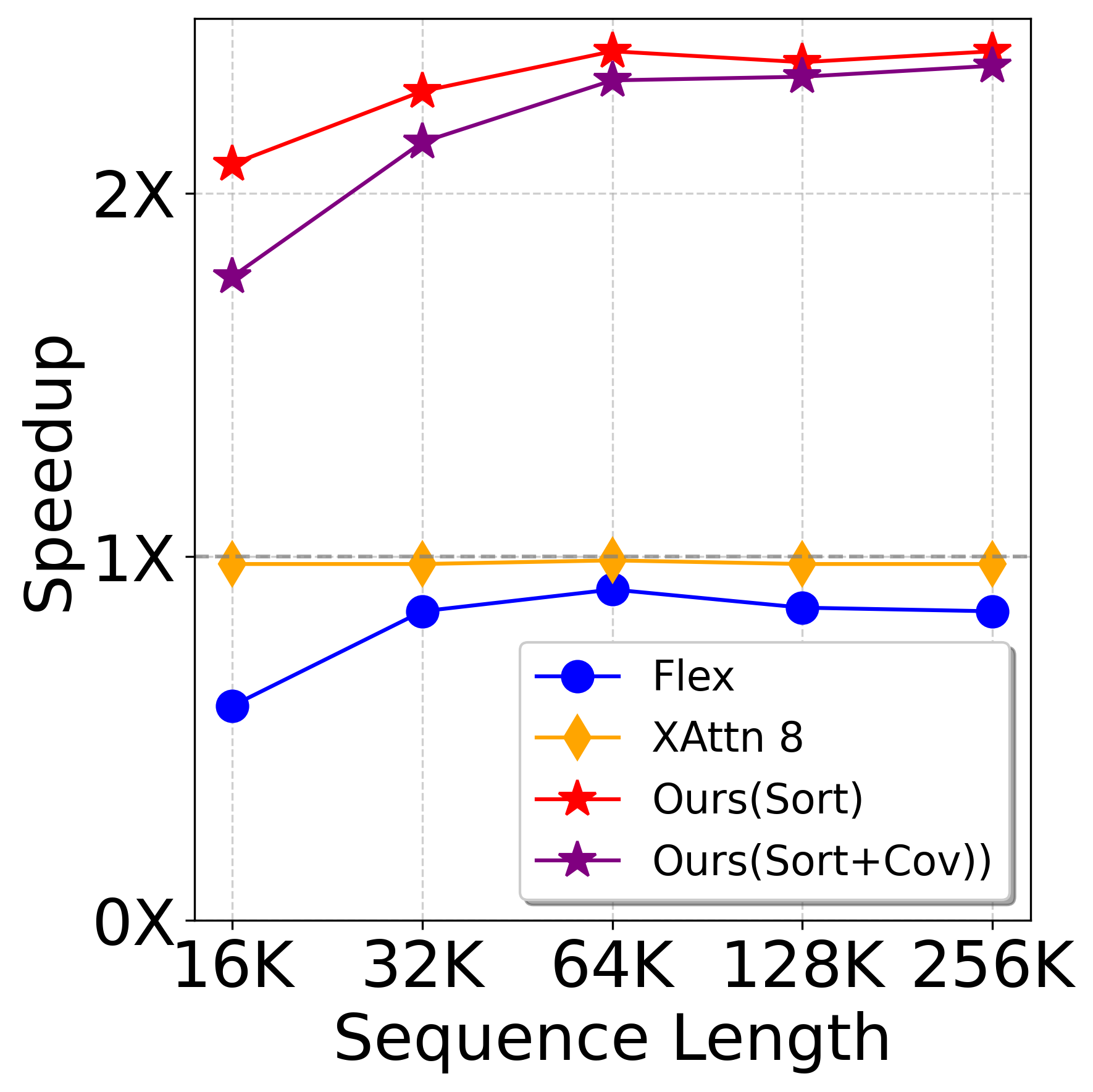}
    \caption{Speedup on RTX3090.}
  \end{subfigure}
  \vspace{-2mm}
  \caption{Operator-level speedup over flash-attention baseline.
        Our method variants consistently outperform prior approaches across various context-length and devices.}
  \label{fig:speedup}
  \vspace{-3mm}
\end{figure}

\paragraph{Attention Acceleration}
We measure operator-level speedup over baseline across sequence lengths from 16K to 256K tokens. As shown in \cref{fig:speedup}, our method variants achieve substantial acceleration, with ``Ours (Sort)'' delivering up to \textbf{6.95$\times$} speedup at 256K significantly outperforming both Flex (4.56$\times$) and XAttn-8 (3.5$\times$). To balance fidelity and efficiency, we apply covariance compensation only in the first and last layers, where attention patterns are most sensitive and diverse. This selective use yields ``Ours (Sort+Cov)'', which retains a strong 5.70$\times$ speedup at 256K while improving output quality. 

\vspace{-3mm}
\paragraph{Norm-based Sort Analysis}

In ~\cref{tab:ruler4k_sparsity}, we present the performance of various norm-based sorting strategies on Ruler-4K under different sparsity levels. 
Notably, jointly sorting queries and keys (SortQ+SortK) further improves performance at higher sparsity levels , suggesting that aligning both query and key orderings enhances the effectiveness of block-sparse attention. 
These results validate that norm-based reordering—particularly of keys—effectively concentrates salient information into fewer blocks, enabling sparse attention to retain more task-relevant signals.

\begin{table}[t]
\small
\setlength{\tabcolsep}{19pt}
\centering
\caption{Performance on Ruler-4K under different norm sort strategies and sparsity settings.}
\vspace{-2mm}
\begin{tabular}{@{}lccc@{}}
\toprule
\textbf{Sparsity} & {90\%} & {70\%} & {50\%} \\
\midrule
Baseline & 22.34 & 52.91 & 81.08 \\
SortQ & 24.07 & 55.66 & 82.64 \\
SortK & \textbf{27.13} & 60.56 & 87.25 \\
SortQ+SortK & 26.98 & \textbf{61.66} & \textbf{88.27} \\
\bottomrule
\end{tabular}
\label{tab:ruler4k_sparsity}
\vspace{-6mm}
\end{table}

\section{Related Work}
\label{sec:related_work}

\paragraph{Diffusion Language Models}
Diffusion-based generative models, originally developed for continuous data like images~\cite{esser2024scalingrectifiedflowtransformers}, have recently been adapted to text generation through continuous~\citep{li2022diffusion,gong2022diffuseq,han2022ssd,strudel2022self,chen2022analog,dieleman2022continuous,richemond2022categorical,wu2023ardiffusion,mahabadi2024tess,ye2023dinoiser,zhang2023planner,lou2023reflected,graves2023bayesian,lin2023text,xue2024unifying, zhang2025target} and 
discrete~\citep{reid2022diffuser,sun2022score,kitouni2023disk,Zheng2023ARD,chen2023fast,ye2023diffusion,gat2024discrete,zheng2024maskeddiffusionmodelssecretly} variants.
Unlike autoregressive models that generate tokens sequentially, Diffusion Language Models (DLMs) denoise an entire sequence in parallel, offering the potential for faster decoding. Recent works such as UltraLLaDA~\cite{ultrallada}, LongLLaDA~\cite{longllada} and LLaDA-V~\cite{lladav} demonstrate that DLMs can achieve competitive performance on language and multi-modal tasks. 
However, the quadratic complexity of full-attention over long denoising sequences remains a major bottleneck, especially in multi-modal or document-level applications.
\vspace{-3mm}
\paragraph{Sparse Attention}
To mitigate the quadratic complexity of self-attention, several sparse attention mechanisms have been proposed~\cite{xi2025sparsevideogenacceleratingvideo,yuan2025nativesparseattentionhardwarealigned,lu2025mobamixtureblockattention,zhang2025fastvideogenerationsliding}. 
Early approaches employ fixed sparsity patterns, such as local windows or strided blocks~\cite{Beltagy2020Longformer, zaheer2020bigbird}, which effectively reduce computational cost but often struggle to capture long-range dependencies crucial for reasoning and understanding global context~\cite{child2019generatinglongsequencessparse,leviathan2024selectiveattentionimprovestransformer}. 
More recent methods adopt adaptive or data-driven sparsity. 
H2O \cite{zhang2023h2o} and TOVA \cite{oren2024transformersmultistaternns} discard tokens based on query patterns. 
SeerAttention~\cite{seeratt} proposes a lightweight learnable module for sparse pattern selection, while XAttention~\cite{xatt} introduces an anti-diagonal block selection strategy that efficiently models long-range interactions. 
StreamingLLM~\cite{attsink} observes that certain early tokens act as “sinks” that stabilize attention in streaming LMs.

\section{Conclusion and Limitation}
\label{sec:conclusion}
\paragraph{Conclusion.}
In this work, we proposed a block-sparse attention framework for diffusion language models, introducing a pre-downsampled block-sparse mechanism that evaluates block informativeness before attention computation. 
Through norm-based ranking and covariance-compensated correction, our method effectively reduces approximation error and achieves substantial acceleration while maintaining near full-attention fidelity across diverse modalities and tasks. 
Extensive experiments demonstrate up to 6.95× speedup on 128K-sequence inference with minimal accuracy degradation, validating the scalability and generalization of our approach in ultra-long context modeling.
\vspace{-3mm}
\paragraph{Limitations.}
Despite these promising results, several limitations remain. First, our current design focuses on diffusion-based models; extending the proposed mechanism to AR models may reveal additional challenges in maintaining causal masking. 
%
Besides, the current covariance-compensated correction involves partial covariance estimation to improve fidelity. There is potential for further acceleration through triton or low-rank approximations.


\section{Acknowledgements}
This work was supported in part by the Research Grants Council under the Areas of Excellence scheme grant AoE/E-601/22-R.

{
    \small
    \bibliographystyle{ieeenat_fullname}
    \bibliography{main}
}



\end{document}


\maketitle
\thispagestyle{empty} 

\appendix 


\section{Proof of some Lemmas}

\begin{lemma}[Normalization perturbation]
\label{thm:normalization_perturbation}
Let $u, v \in \mathbb{R}^n_{>0}$ and denote
\[
\alpha = \sum_{k=1}^n u_k,
\qquad
\beta  = \sum_{k=1}^n v_k.
\]
Define the normalized vectors
\[
\pi(u) = \frac{u}{\alpha},
\qquad
\pi(v) = \frac{v}{\beta}.
\]
Then
\begin{equation}
    \big\|\pi(u) - \pi(v)\big\|_1
    \le
    \frac{4}{\alpha + \beta}\,\|u - v\|_1.
    \label{eq:normalization_perturbation}
\end{equation}
In particular, when $\alpha$ and $\beta$ are of the same order,
the $\ell_1$ distance between the normalized vectors is controlled
(up to a constant factor) by the $\ell_1$ distance between the
unnormalized vectors.
\end{lemma}

\begin{proof}
For each coordinate $k$,
\[
\pi(u)_k - \pi(v)_k
= \frac{u_k}{\alpha} - \frac{v_k}{\beta}
= \frac{u_k \beta - v_k \alpha}{\alpha \beta}
= \frac{\beta(u_k - v_k) + v_k(\beta - \alpha)}{\alpha \beta}.
\]
Taking absolute values and summing over $k$ yields
\[
\big\|\pi(u) - \pi(v)\big\|_1
\le
\frac{1}{\alpha \beta}
\Big(
  \beta \sum_{k=1}^n |u_k - v_k|
  + |\beta - \alpha|\sum_{k=1}^n v_k
\Big)
=
\frac{1}{\alpha \beta}
\Big(
  \beta \|u - v\|_1
  + \beta\,|\beta - \alpha|
\Big).
\]
Since
\[
|\beta - \alpha|
= \Big|\sum_{k=1}^n (v_k - u_k)\Big|
\le \sum_{k=1}^n |u_k - v_k|
= \|u - v\|_1,
\]
we obtain
\[
\big\|\pi(u) - \pi(v)\big\|_1
\le
\frac{2\beta}{\alpha \beta} \,\|u - v\|_1
=
\frac{2}{\alpha}\,\|u - v\|_1.
\]
By symmetry (exchanging $u$ and $v$), we also have
$\|\pi(u) - \pi(v)\|_1 \le \frac{2}{\beta}\,\|u - v\|_1$.
Therefore,
\[
\big\|\pi(u) - \pi(v)\big\|_1
\le
2 \min\Big\{\frac{1}{\alpha}, \frac{1}{\beta}\Big\}\,\|u - v\|_1
\le
\frac{4}{\alpha + \beta}\,\|u - v\|_1,
\]
where we used
$\min\{\alpha,\beta\} \le (\alpha + \beta)/2$ and thus
$\min\{1/\alpha,1/\beta\} \le 2/(\alpha + \beta)$.
This proves \eqref{eq:normalization_perturbation}.
\end{proof}